\documentclass[11pt,a4paper]{article}
\usepackage[hyperref]{acl2020}
\usepackage{times}
\usepackage{latexsym}

\usepackage{microtype}

\usepackage{makecell}
\usepackage{multirow}
\usepackage{amsmath}
\usepackage{amssymb}
\usepackage{enumitem}
\usepackage{color}
\usepackage{graphicx}
\usepackage{comment}
\usepackage{float}

\aclfinalcopy 
\def\aclpaperid{1038} 

\title{SongNet: Rigid Formats Controlled Text Generation}

\author{Piji Li \ \ Haisong Zhang \ \ Xiaojiang Liu \ \ Shuming Shi\\
Tencent AI Lab, Shenzhen, China\\
\texttt{\{pijili,hansonzhang,kieranliu,shumingshi\}@tencent.com}
}

\date{}

\begin{document}
\maketitle
\begin{abstract}
Neural text generation has made tremendous progress in various tasks. One common characteristic of most of the tasks is that the texts are not restricted to some rigid formats when generating. However, we may confront some special text paradigms such as Lyrics (assume the music score is given), Sonnet, SongCi (classical Chinese poetry of the Song dynasty), etc. The typical characteristics of these texts are in three folds: (1) They must comply fully with the rigid predefined formats. (2) They must obey some rhyming schemes. (3) Although they are restricted to some formats, the sentence integrity must be guaranteed. To the best of our knowledge, text generation based on the predefined rigid formats has not been well investigated. Therefore, we propose a simple and elegant framework named \textbf{SongNet} to tackle this problem. The backbone of the framework is a Transformer-based auto-regressive language model. Sets of symbols are tailor-designed to improve the modeling performance especially on format, rhyme, and sentence integrity. We improve the attention mechanism to impel the model to capture some future information on the format. A pre-training and fine-tuning framework is designed to further improve the generation quality.
Extensive experiments conducted on two collected corpora demonstrate that our proposed framework generates significantly better results in terms of both automatic metrics and the human evaluation.\footnote{Code: \url{http://github.com/lipiji/SongNet}}
\end{abstract}

\section{Introduction}
Recent years have seen the tremendous progress in the area of natural language generation especially benefiting by the neural network models such as Recurrent Neural Networks (RNN) or Convolutional Neural Networks (CNN) based sequence-to-sequence (seq2seq) frameworks \cite{bahdanau2014neural,gehring2017convolutional}, Transformer and its variants \cite{vaswani2017attention,dai2019transformer}, pre-trained auto-regressive language models such as XLNet \cite{yang2019xlnet} and GPT2 \cite{radford2019language}, etc. 
Performance has been improved significantly in lots of tasks such as machine translation \cite{bahdanau2014neural,vaswani2017attention}, dialogue systems \cite{vinyals2015neural,shang2015neural,li2020empirical}, text summarization \cite{rush2015neural,li2017deep,see2017get}, story telling \cite{fan2018hierarchical,see2019massively}, poetry writing \cite{zhang2014chinese,lau2018deep,liao2019gpt}, etc.

\begin{figure}[t!]
\centering
\includegraphics[width=0.99\columnwidth]{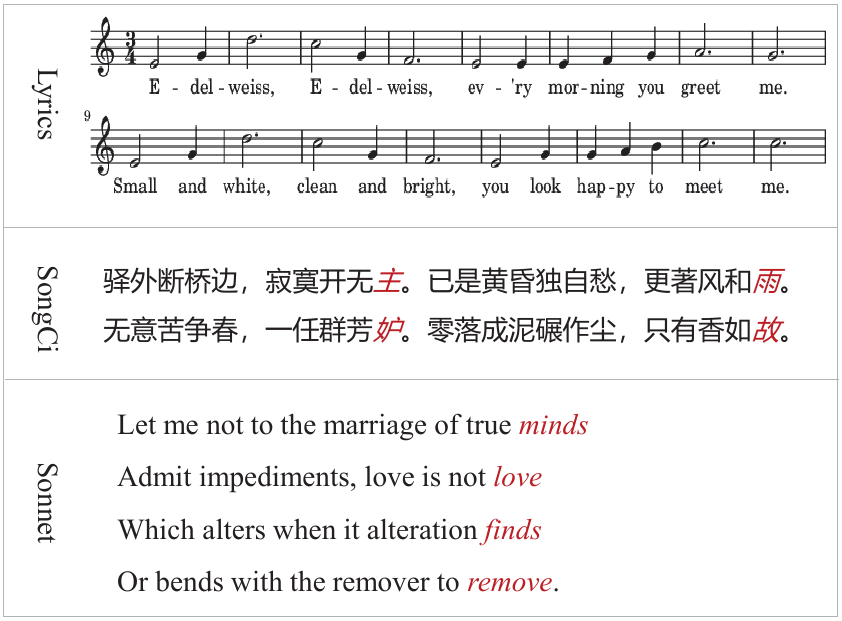}
\caption{Examples of text with rigid formats. In lyrics, the syllables of the lyric words must align with the tones of the notation. In SongCi and Sonnet, there are strict rhyming schemes and the rhyming words are labeled in red color and \textit{italic} font.}
\label{fig:front}
\end{figure}

Generally, most of the above mentioned tasks can be regarded as free text generation, which means that no constraints on the format and structure, say the number of words and rhyming rules. Note that tasks of dialogue generation and story telling are almost in an open-ending generation style as long as the generated content is relevant with the conditional input text. 
Although there are formats constraints on the poetry text, the proposed models just treat the formats as kind of latent information and let the model capture this feature implicitly during training \cite{liao2019gpt}. The model trained on the five-character quatrain corpus cannot generate seven-character verses.
Moreover, it is impossible to trigger these models to generate satisfying results according to arbitrary new defined formats. 

In practice we will confront some special text paradigms such as Lyrics (assume the music score is given), Sonnet (say Shakespeare's Sonnets \cite{shakespeare2000shakespeare}), SongCi (a kind of Ci. Ci is a type of lyric poetry in the tradition of Classical Chinese poetry.\footnote{http://en.wikipedia.org/wiki/Ci\_(poetry)}, SongCi is the Ci created during Song dynasty), etc., and some examples are illustrated in Figure~\ref{fig:front}. The typical characteristics of these text can be categorized into three folds: (1) The assembling of text must comply fully with the predefined rigid \textbf{formats}. Assume that the music score is composed, then the lyricist must fill the lyric content strictly tally with the schemes lie in the notation. Take partial of song ``Edelweiss'' as shown in the first row of Figure~\ref{fig:front} as example, the syllables of the lyric words must align with the tones of the notation. The second row of Figure~\ref{fig:front} depicts the content of a SongCi created based on the CiPai of ``Bu Suan Zi''. Given the CiPai, the number of characters and the syntactical structure of the content are also defined (e.g., the number of characters of each clause: 5, 5. 7, 5. 5, 5. 7, 5.).
(2) The arrangement of the content must obey the defined \textbf{rhyming} schemes. For example, all the final words (words in red color and \textit{italic} font) of the SongCi content in Figure\ref{fig:front} are rhyming (the spelling of each word is: ``zhu'', ``yu'', ``du'', and ``gu''.). The example in the third row of Figure~\ref{fig:front} comes from Shakespeare's ``Sonnet 116'' \cite{shakespeare2000shakespeare}, the first four sentences. Usually, the rhyming schemes of Shakespeare's Sonnets is ``\textit{ABAB CDCD EFEF GG}'' \footnote{http://en.wikipedia.org/wiki/Shakespeare\%27s\_sonnets}. In the example, the rhyming words in scheme ``ABAB'' are ``minds'', ``love'', ``finds'', and ``remove''. (3) Even though the format is rigid, the sentence \textbf{integrity} must always be guaranteed. Incomplete sentence such as ``love is not the'' is inappropriate.

To the best of our knowledge, text generation based on the predefined rigid formats constraints has not been well investigated yet. In this work, we propose a simple and elegant framework named \textbf{SongNet} to address this challenging problem. The backbone of the framework is a Transformer-based auto-regressive language model. Considering the three folds characteristics mentioned above, we introduce sets of tailor-designed indicating symbols to improve the modeling performance, especially for the robustness of the format, rhyme, as well as sentence integrity. We improve the attention mechanism to impel the model to capture the future information on the format to further enhance sentence integrity. Inspired by BERT \cite{devlin2019bert} and GPT \cite{radford2018improving,radford2019language}, a pre-training and fine-tuning framework is designed to further improve the generation quality. To verify the performance of our framework, we collect two corpora, SongCi and Sonnet, in Chinese and English respectively. Extensive experiments on the collected datasets demonstrate that our proposed framework can generate satisfying results in terms of both the tailor-designed automatic metrics including format accuracy, rhyming accuracy, sentence integrity, as well as the human evaluation results on relevance, fluency, and style.

\begin{figure*}[!t]
\centering
\includegraphics[width=1.99\columnwidth,height=5.5cm]{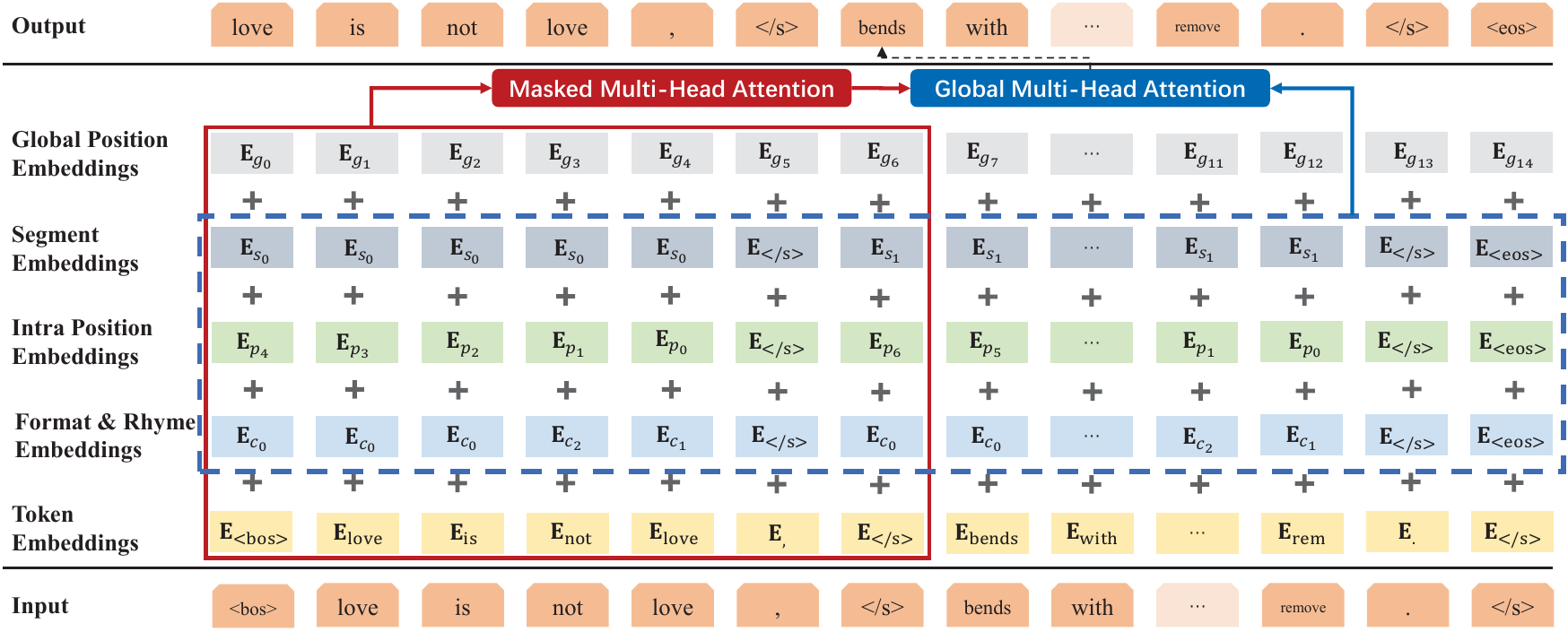}
\caption{The framework of our proposed model.}
\label{fig:framework}
\end{figure*}

In summary, our contributions are as follows:
\begin{itemize}[topsep=0pt]
    \setlength\itemsep{-0.5em}
    \item We propose to tackle a new challenging task: rigid formats controlled text generation. A pre-training and fine-tuning framework named SongNet is designed to address the problem.
    \item Sets of symbols are tailor-designed to improve the modeling performance. We improve the attention mechanism to impel the model to capture the future information to further enhance the sentence integrity.
    \item To verify the performance of our framework SongNet, we collect two corpora, SongCi and Sonnet, in Chinese and English respectively. We design several automatic evaluation metrics and human evaluation metrics to conduct the performance evaluation.
    \item Extensive experiments conducted on two collected corpora demonstrate that our proposed framework generates significantly better results given arbitrary formats, including the cold-start formats or even the formats newly defined by ourselves.
\end{itemize}

\section{Task Definition}
\label{sec:task}

The task of rigid formats controlled text generation is defined as follows:

\noindent \textbf{Input}: a rigid format $C \in \mathcal{C}$:
\begin{equation}
    C = \{c_0\ c_1\ c_2\ c_3,\ c_0\ c_1\ c_2\ c_3\ c_4\ c_5.\} 
\end{equation}
where $\mathcal{C}$ is the set of all possible formats. Note that we can define arbitrary new formats not restricted to the ones pre-defined in the corpus, thus $|\mathcal{C}| \to \infty$.  Format token $c_i$ denotes a place-holder symbol of $C$ which need to be translated into a real word token. Format $C$ contains $10$ words plus two extra punctuation characters ``,'' and ``.''

\noindent \textbf{Output}: a natural language sentence $Y \in \mathcal{Y}$ which tally with the defined format $C$:
\[
\begin{aligned}
Y =\ &love\ is\ not\ love, \\
&bends\ with\ the\ remover\ to\ remove.
\end{aligned}
\]
where the example sentences are extracted from the Shakespeare's Sonnets \cite{shakespeare2000shakespeare}. From the result $Y$ we can observe that the count of words is 10 which is consistent with the format $C$. The punctuation characters ``,'' and ``.'' are also correct. Thus, we claim that it is a $100\%$ format accuracy result. Also, since the two clause sentences are complete, we can get a good sentence integrity score. If $\mathcal{C}$ is defined on the literary genres of SongCi or Sonnet which have rhyming constraints, the rhyming performance should be evaluated as well. Recall that $\mathcal{C}$ can be arbitrary and flexible, thus we can rebuild a new format $C'$ based on the generated result $Y$ by masking partial content, say $ C' = \{c_0\ c_1\ c_2\ \ love,\ c_0\ c_1\ c_2\ c_3\ c_4\ \ remove.\}$, then we may obtain better results by re-generating based on $C'$. We name this operation as \textbf{polishing}. 

Finally, the target of this problem is to find a mapping function $G$ to conduct the rigid formats controlled text generation:
\begin{equation}
    Y = G(C)
\end{equation}

\section{Framework Description}

\subsection{Overview}

As shown in Figure~\ref{fig:framework}, the backbone of our framework is a Transformer-based auto-regressive language model. The input can be the whole token sequences of samples from SongCi or Sonnet. We tailor-design several sets of indicating symbols to enhance the performance in terms of accuracy on format, rhyme, and sentence integrity. Specifically, symbols $C = \{c_i\}$ are introduced for format and rhyming modeling; Intra-position symbols $P = \{p_i\}$ are designed to represent the local positions of the tokens within each sentence aiming to improve the rhyming performance and the sentence integrity. Segment symbols $S = \{s_i\}$ are employed to identify the sentence border to further improve the sentence quality.
Attention mechanism is improved to impel the model to capture the future format information such as the sentence ending markers. Similar to BERT \cite{devlin2019bert} and GPT \cite{radford2018improving,radford2019language}, pre-training and fine-tuning paradigm is utilized to boost the performance of the original models.

\subsection{Details}
We use two sentences (as shown in Figure~\ref{fig:front}) ``love is not love, ..., bends with the remover to remove'' extracted from the Shakespeare's Sonnets \cite{shakespeare2000shakespeare} as examples to describe the details of our framework SongNet. Since our basic model is a Transformer-based auto-regressive language model, during training, the input is ``$\langle bos \rangle$ love is not love, $\langle /s \rangle$ ..., bends with the remover to remove. $\langle /s \rangle$'', and the corresponding output is a left-shifting version of the input (tokenized, and we ignore ``...'' for convenience and clarity):
\[
\small
\begin{aligned}
&love\ is\ not\ love\ ,\ \langle /s \rangle \\
&bends\ with\ the\ remover\ to\ remove\ .\ \langle /s \rangle\ \langle eos \rangle
\end{aligned}
\]
where $\langle /s \rangle$ denotes the clause or sentence separator, and $\langle eos \rangle$ is the ending marker of the whole sequence.
The target of our framework is to conduct the formats controlled text generation. Therefore, the indicating symbols for format and rhyme as well as the sentence integrity are designed based on the target output sequence. 

\noindent \textbf{Format and Rhyme Symbols}:
\begin{equation}
\begin{aligned}
    C = \{&c_0, c_0, c_0, c_2, c_1, \langle /s \rangle \\
     &c_0, c_0, c_0, c_0, c_0, c_2, c_1, \langle /s \rangle, \langle eos \rangle
        \}
\end{aligned}
\end{equation}
where we use $\{c_0\}$ to represent the general tokens; $\{c_1\}$ depict the punctuation characters; $\{c_2\}$ represent the rhyming tokens ``love'' and ``remove''. $\langle /s \rangle$  and $\langle eos \rangle$ are kept.

\noindent \textbf{Intra-Position Symbols}:
\begin{equation}
\begin{aligned}
    P = \{&p_4, p_3, p_2, p_1, p_0, \langle /s \rangle \\
     &p_6, p_5, p_4, p_3, p_2, p_1, p_0, \langle /s \rangle, \langle eos \rangle
        \}
\end{aligned}
\end{equation}
$\{p_i\}$ denote the local positions of tokens within the same clause or sentence. Note that we align the position symbol indices in a \textbf{descending order}. The aim is to improve the sentence integrity by impelling the symbols capture the sentence dynamic information, precisely, the sense to end a sequence. For example, $\{p_0\}$ usually denote punctuation characters, thus $\{p_1\}$ should be the ending words of sentences. 

\noindent \textbf{Segment Symbols}:
\begin{equation}
\begin{aligned}
    S = \{&s_0, s_0, s_0, s_0, s_0, \langle /s \rangle \\
     &s_1, s_1, s_1, s_1, s_1, s_1, s_1, \langle /s \rangle, \langle eos \rangle
        \}
\end{aligned}
\end{equation}
where $s_i$ is the symbol index for sentence $i$. The purpose is to enhance the interactions between different sentences in different positions by defining the sentence index features.

During training, all the symbols as well as the input tokens are fed into the transformer-based language model. Contrast to Transformer \cite{vaswani2017attention}, BERT \cite{devlin2019bert}, and GPT2 \cite{radford2019language}, we modify the traditional attention strategies slightly to fit our problem. 

Specifically, for the input, we first obtain the representations by summing all the embeddings of the input tokens and symbols, as shown in the red solid box of Figure~\ref{fig:framework}:
\begin{equation}
    \mathbf{H}^0_t = \mathbf{E}_{w_t}+\mathbf{E}_{c_t}+\mathbf{E}_{p_t}+\mathbf{E}_{s_t}+\mathbf{E}_{g_t}
\end{equation}
where $0$ is the layer index and  $t$ is the state index. $\mathbf{E}_*$ is the embedding vector for input $*$. $w_t$ is the real token at position $t$. $c$, $p$, and $s$ are three pre-defined symbols. $g$ is the global position index same as position symbols used in Transformer \cite{vaswani2017attention}.

Moreover, the state at time $t$ need to know some future information to grasp the global sequence dynamic information. For example, the model may want to know if it should close the decoding progress by generating the last word and a punctuation character to end the sentence. To represent the global dynamic information, we introduce another variable $\mathbf{F}^0$ by only summing the pre-defined symbols as shown in the blue dash box of Figure~\ref{fig:framework}:
\begin{equation}
    \mathbf{F}^0_t = \mathbf{E}_{c_t}+\mathbf{E}_{p_t}+\mathbf{E}_{s_t}
    \label{eql:formant_f}
\end{equation}

After processing the input, two blocks of attention mechanisms are introduced to conduct the feature learning procedure. The first block is a masking multi-head self-attention component, and the second block is named global multi-head attention.

\noindent \textbf{Masking Multi-Head Self-Attention}:
\begin{equation}
\begin{split}
    \mathrm{\bf C}^{1}_{t} &= \textsc{Ln}\left(\textsc{Ffn} (\mathrm{\bf C}^{1}_{t}) +\mathrm{\bf C}^1_t \right) \\
    \mathrm{\bf C}^{1}_{t} &= \textsc{Ln}\left(\textsc{Slf-Att} (\mathrm{\bf Q}^{0}_{t}, \mathrm{\bf K}^{0}_{\leq t}, \mathrm{\bf V}^{0}_{\leq t}) +\mathrm{\bf H}^0_t \right) \\
    \mathrm{\bf Q}^{0} &=  \mathrm{\bf H}^{0} \mathrm{\bf W}^{Q} \\
    \mathrm{\bf K}^{0}, \mathrm{\bf V}^{0} &= \mathrm{\bf H}^{0}\mathrm{\bf W}^{K}, \mathrm{\bf H}^{0} \mathrm{\bf W}^{V}
\end{split}
\label{eql:formant_c1}
\end{equation}
where \textsc{Slf-Att}($\cdot$), \textsc{Ln}($\cdot$), and \textsc{Ffn}($\cdot$) represent self-attention mechanism, layer normalization, and feed-forward network respectively. Note that we only use the states whose indices $\leq t$ as the attention context.

After obtaining $\mathbf{C_t^1}$ from Equation~(\ref{eql:formant_c1}), we feed it into the second attention block to capture the global dynamic information from $\mathbf{F^0}$.

\noindent \textbf{Global Multi-Head Attention}:
\begin{equation}
\begin{split}
    \mathrm{\bf H}^{1}_{t} &= \textsc{Ln}\left(\textsc{Ffn} (\mathrm{\bf H}^{1}_{t}) +\mathrm{\bf H}^1_t \right) \\
    \mathrm{\bf H}^{1}_{t} &= \textsc{Ln}\left(\textsc{\small{Global-Att}} (\mathrm{\bf Q}^{1}_{t}, \mathrm{\bf K}^{1}, \mathrm{\bf V}^{1}) +\mathrm{\bf C}^1_t \right) \\
    \mathrm{\bf Q}^{1} &=  \mathrm{\bf C}^{1} \mathrm{\bf W}^{Q} \\
    \mathrm{\bf K}^{1}, \mathrm{\bf V}^{1} &= \mathrm{\bf F}^{0}\mathrm{\bf W}^{K}, \mathrm{\bf F}^{0} \mathrm{\bf W}^{V}
\end{split}
\end{equation}
We can observe that all the context information from $\mathrm{\bf F}^{0}$ are considered. This is the reason why we name it as ``global attention'' and why the input real token information $\mathbf{E}_{w_t}$ is NOT considered.
Then the calculation of the unified first model layer is finished. We can iteratively apply these two attention blocks on the whole $L$ model layers until obtain the final representations $\mathbf{H}^L$. Note that $\mathbf{H}$ is renewed layerly, however the global variable $\mathbf{F}^0$ is fixed.

Finally, the training objective is to minimize the negative log-likelihood over the whole sequence:
\begin{equation}
\begin{split}
    \mathcal{L}^{\mathrm{nll}} = -\sum^{n}_{t=1} \log P(\mathrm{\bf y}_t|\mathrm{\bf y}_{<t})
\end{split}
\label{eql:nll}
\end{equation}

\subsection{Pre-training and Fine-tuning}
Although our framework can be trained purely on the training dataset of the target corpus, usually the scale of the corpus is limited. For example, there are only about 150 samples in the corpus of Shakespeare's Sonnets \cite{shakespeare2000shakespeare}. Therefore, we also design a pre-training and fine-tuning framework to further improve the generation quality. 

Recall that in the task definition in Section~\ref{sec:task}, we claim that our model owns the ability of refining and polishing. To achieve this goal, we adjust the masking strategy used in BERT \cite{devlin2019bert} to our framework according to our definitions. Specifically, we randomly  (say 20\%) select partial of the original content and keep them not changed when building the format symbols $C$. For example, we will get a new symbol set $C'$ for the example sentences:
\[
\small
\begin{aligned}
    C' = \{&c_0, c_0, c_0, love, c_1, \langle /s \rangle \\
     &bends, c_0, c_0, c_0, c_0, remove, c_1, \langle /s \rangle, \langle eos \rangle
        \}
\end{aligned}
\]
where ``love'', ``bends'' and ``remove'' are kept in  the format $C'$.

After the pre-training stage, we can conduct the fine-tuning procedure directly on the target corpus without adjusting any model structure. 

\subsection{Generation}
We can assign any format and rhyming symbols $C$ to control the generation. Given $C$, we will obtain $P$ and $S$ automatically. And the model can conduct generation starting from the special token $\langle bos \rangle$ iteratively until meet the ending marker $\langle eos \rangle$. Both beam-search algorithm \cite{koehn2004pharaoh} and truncated top-k sampling \cite{fan2018hierarchical,radford2019language} method are utilized to conduct the decoding.

\begin{table*}[hbt!]
\centering
\resizebox{1.75\columnwidth}{!}{
\begin{tabular}{l|cc|cc|cc}
\Xhline{2\arrayrulewidth}
\multirow{2}{*}{\textbf{Model}} & \multicolumn{2}{c|}{\textbf{PPL$\downarrow$}} & \multicolumn{4}{c}{\textbf{Diversity (Distinct) $\uparrow$}}\\
\cline{2-7}& \textsc{Val} & \textsc{Test} & \textsc{Ma-D-1} & \textsc{Mi-D-1} & \textsc{Ma-D-2} & \textsc{Mi-D-2}\\
\hline
\hline
S2S & 19.61 & 20.43  &75.35 & 2.48 &98.35 & 36.23 \\
GPT2& 148.11 & 104.99 & - & - & - &- \\
GPT2 w/ Fine-tuning & 18.25 & 17.00 & 73.87 & 2.57 & 96.07 & 33.92  \\
SongNet (only Pre-training) & 24.41 & 16.23 & 74.84 & 4.59 & 95.09 & 54.98 \\
SongNet (only Fine-tuning) & 12.75 & 14.73 & 75.96 & 2.69 & 97.59 & 37.26 \\
SongNet & \textbf{11.56} & \textbf{12.64} & 75.04 & 2.66 & 97.29 & 36.78\\
\Xhline{2\arrayrulewidth}
\end{tabular}}
\resizebox{1.75\columnwidth}{!}{
\begin{tabular}{l|cc|cc|c}
\Xhline{2\arrayrulewidth}
\multirow{2}{*}{\textbf{Model}} & \multicolumn{2}{c|}{\textbf{Format$\uparrow$}} & \multicolumn{2}{c|}{\textbf{Rhyme$\uparrow$}} & \multirow{2}{*}{\textbf{Integrity$\downarrow$}}\\
\cline{2-5} & \textsc{Ma-F1} & \textsc{Mi-F1} & \textsc{Ma-F1} & \textsc{Mi-F1}\\
\hline
\hline
S2S & 44.32 & 38.16  & 53.80 & 52.27 & 8.30$\pm$2.06\\
GPT2 w/ Fine-tuning & 35.70 &35.20 & 53.48&52.50 & 45.92$\pm$20.12  \\
SongNet (only Pre-training) & 29.12 & 29.46 & 53.77 & 53.13 & 30.98$\pm$14.06\\
SongNet (only Fine-tuning) & 99.81 & 99.83 & \textbf{79.23} & \textbf{78.63} & 2.14$\pm$0.10 \\
SongNet & \textbf{99.88} & \textbf{99.89} & 73.21 & 72.59 & \textbf{1.77$\pm$0.16} \\
\Xhline{2\arrayrulewidth}
\end{tabular}}
\caption{Automatic evaluation results on SongCi}
\label{tbl:exp-songci}
\end{table*}

\begin{table*}[hbt!]
\centering
\resizebox{1.75\columnwidth}{!}{
\begin{tabular}{l|cc|cc|cc}
\Xhline{2\arrayrulewidth}
\multirow{2}{*}{\textbf{Model}} & \multicolumn{2}{c|}{\textbf{PPL$\downarrow$}} & \multicolumn{4}{c}{\textbf{Diversity (Distinct) $\uparrow$}}\\
\cline{2-7}& \textsc{Val} & \textsc{Test} & \textsc{Ma-D-1} & \textsc{Mi-D-1} & \textsc{Ma-D-2} & \textsc{Mi-D-2}\\
\hline
\hline
GPT2 w/ Fine-tuning & 31.47 & 31.03 & 73.87 & 2.57 & 96.07 & 33.92  \\
SongNet (only Pre-training) & 28.56 & 28.07 & 49.92 & 25.14 & 85.35 & 65.70 \\
SongNet (only Fine-tuning) & 34.62 & 34.53 & 42.31 & 4.96 & 90.76 & 47.26 \\
SongNet & \textbf{27.46} & \textbf{27.63} & 43.01 & 10.43 & 80.06 & 56.14\\
\Xhline{2\arrayrulewidth}
\end{tabular}}
\resizebox{1.75\columnwidth}{!}{
\begin{tabular}{l|cc|cc|c}
\Xhline{2\arrayrulewidth}
\multirow{2}{*}{\textbf{Model}} & \multicolumn{2}{c|}{\textbf{Format$\uparrow$}} & \multicolumn{2}{c|}{\textbf{Rhyme$\uparrow$}} & \multirow{2}{*}{\textbf{Integrity$\downarrow$}}\\
\cline{2-5} & \textsc{Ma-F1} & \textsc{Mi-F1} & \textsc{Ma-F1} & \textsc{Mi-F1}\\
\hline
\hline
GPT2 w/ Fine-tuning & 2.03 &1.91 & 5.20 & 6.24 & 15.77$\pm$3.63  \\
SongNet (only Pre-training) & 99.99 & 99.99 & 3.93 & 4.01 & 15.28$\pm$2.04\\
SongNet (only Fine-tuning) & 99.25 & 99.99 & 7.50 & 7.41 & 18.86$\pm$2.59 \\
SongNet & 98.73 & 98.73 & \textbf{11.46} & \textbf{11.41} & \textbf{11.86$\pm$3.01} \\
\Xhline{2\arrayrulewidth}
\end{tabular}}
\caption{Automatic evaluation results on Sonnet}
\label{tbl:exp-sonnet}
\end{table*}

\section{Experimental Setup}

\subsection{Settings}
The parameter size of our model are fixed in both the pre-training stage and the fine-tuning stage. The number of layers $L=12$, and hidden size is 768. We employ 12 heads in both the masking multi-head self-attention block and the global attention block. Adam \cite{kingma2014adam} optimization method with Noam learning-rate decay strategy and 10,000 warmup steps is employed to conduct the pre-training.

\subsection{Datasets}

\begin{table}[!t]
\centering
\begin{tabular}{l|c|c|c|c}
\Xhline{2\arrayrulewidth}
 Corpus & \#Train & \#Dev & \#Test & \#Vocab \\
 \hline
 SongCi & 19,244 & 847 &  962 & 5310  \\
 \hline
 Sonnet & 100 & 27 & 27 & 2801    \\
 \Xhline{2\arrayrulewidth}
\end{tabular}
\caption{Statistics of the datasets SongCi and Sonnet.}
\label{tab:datasets}
\end{table}

We conduct all the experiments on two collected corpus with different literary genres: SongCi and Sonnet, in Chinese and English respectively. The statistic number are shown in Table~\ref{tab:datasets}. We can see that Sonnet is in small size since we only utilize the samples from the Shakespeare's Sonnets \cite{shakespeare2000shakespeare}.
Since SongCi and Sonnet are in different languages, thus we conduct the pre-training procedure on two large scale corpus in the corresponding languages respectively. For Chinese, we collect Chinese Wikipedia (1700M Characters) and a merged Chinese News (9200M Characters) corpus from the Internet. We did not conduct the word segmenting operations on the Chinese datasets, which means that we just use the characters to build the vocabulary, and the size is 27681.
For English, same as BERT, we employ English Wikipedia (2400M words) and BooksCorpus (980M words) \cite{zhu2015aligning} to conduct the pre-training. We did not use BPE operation \cite{sennrich2015neural} on this corpus considering the format controlling purpose. We keep the most frequent 50,000 words to build the vocabulary.

\begin{table*}[!t]
\centering
\resizebox{1.7\columnwidth}{!}{
\begin{tabular}{l|cc|cc|cc}
\Xhline{2\arrayrulewidth}
\multirow{2}{*}{\textbf{Model}} & \multicolumn{2}{c|}{\textbf{PPL$\downarrow$}} & \multicolumn{4}{c}{\textbf{Diversity (Distinct) $\uparrow$}}\\
\cline{2-7}& \textsc{Val} & \textsc{Test} & \textsc{Ma-D-1} & \textsc{Mi-D-1} & \textsc{Ma-D-2} & \textsc{Mi-D-2}\\
\hline
\hline
SongNet & \textbf{12.75} & \textbf{14.73} & 75.96 & 2.69 & 97.59 & 37.26 \\
SongNet-GRU & 16.52 & 20.49 & 74.73 & 1.77 & 98.30 & 28.98\\
SongNet w/o C & 13.51 & 15.38 & 75.42 & 2.48 & 97.36 & 34.85 \\
SongNet w/o P & 14.16 & 17.16 & 73.73 & 2.56 & 97.52 & 34.82 \\
SongNet w/ inverse-P & 13.40 & 15.13 & 74.95 & 2.54 & 97.76 & 35.65 \\
SongNet w/o S & 13.23 & 15.44 & 75.38 & 2.74 & 97.31 & 37.50\\
\Xhline{2\arrayrulewidth}
\end{tabular}}
\resizebox{1.7\columnwidth}{!}{
\begin{tabular}{l|cc|cc|c}
\Xhline{2\arrayrulewidth}
\multirow{2}{*}{\textbf{Model}} & \multicolumn{2}{c|}{\textbf{Format$\uparrow$}} & \multicolumn{2}{c|}{\textbf{Rhyme$\uparrow$}} & \multirow{2}{*}{\textbf{Integrity$\downarrow$}}\\
\cline{2-5} & \textsc{Ma-F1} & \textsc{Mi-F1} & \textsc{Ma-F1} & \textsc{Mi-F1}\\
\hline
\hline
SongNet & 99.81 & 99.83 & 79.23 &78.63 & 2.14$\pm$0.10 \\
SongNet-GRU & 98.99 & 98.99 & 52.13 & 50.93  & 3.28$\pm$1.67\\
SongNet w/o C & 84.73 & 85.39 & 78.59 & 78.24 &  1.77$\pm$0.53\\
SongNet w/o P & 99.61 & 99.59 & 67.85 & 67.29 &  3.33$\pm$0.18\\
SongNet w/ inverse-P & 99.68 & 99.69 & 65.89 & 65.43 &  2.24$\pm$0.21 \\
SongNet w/o S & 99.84 & 99.86 & 80.43 & 80.13 &  1.99$\pm$0.10\\
\Xhline{2\arrayrulewidth}
\end{tabular}}
\caption{Ablation analysis on SongCi}
\label{tbl:ablation-songci}
\end{table*}

\subsection{Evaluation Metrics}
Besides \textbf{PPL} and \textbf{Distinct} \cite{li2016diversity}, we also tailor-design several metrics for our task to conduct the evaluation for format, rhyme, and sentence integrity. 

\noindent \textbf{Format}
Assume that there are $m$ sentences defined in the format $C = \{C^s_1, C^s_2, ..., C^s_m\}$, and the generated results $Y$ contains $n$ sentences $Y = \{Y^s_1, Y^s_2, ..., Y^s_n\}$. Without loss of generality, we align $C$ and $Y$ from the beginning, and calculate the format quality according to the following rules: (1) the length difference $||C^s_i|-|Y^s_i|| \leq \delta$; (2) the punctuation characters must be same.
For SongCi, we let $\delta=0$ and rule (2) must be conforming. For Sonnet, we relax the condition where we let $\delta=1$ and ignore rule (2). Assume that the number of format-correct sentences is $n'$, then we can obtain Precision $p = n'/n$, Recall $r = n'/m$, and F1-measure. We report both the \textbf{Macro-F1} and \textbf{Micro-F1} in the results tables.

\noindent \textbf{Rhyme}
For SongCi, usually, there is only one group of rhyming words in one sample. As the example shown in Table~\ref{fig:front}, the pronunciation of the red rhyming words are ``zhu'', ``y\"u'', ``du'', and ``gu'' respectively, and the rhyming phoneme is ``u''. For the generated samples, we first use the tool pinyin\footnote{http://github.com/mozillazg/python-pinyin} to get the pronunciations (PinYin) of the words in the rhyming positions, and then conduct the evaluation.  
For Shakespeare's Sonnets corpus, the rhyming rule is clear ``\textit{ABAB CDCD EFEF GG}'' and there are 7 groups of rhyming tokens. For the generated samples, we employ the CMU Pronouncing Dictionary\footnote{http://www.speech.cs.cmu.edu/cgi-bin/cmudict} \cite{dictionary1998carnegie} to obtain the phonemes of the words in the rhyming positions. 
For example, the phonemes for word ``asleep'' and ``steep'' are ['AH0', 'S', 'L', 'IY1', 'P'] and ['S', 'T', 'IY1', 'P'] respectively. 
And then we can conduct the evaluation by counting the overlapping units from both the original words and the extracted phonemes group by group. We report the Macro-F1 and Micro-F1 numbers in the results tables as well.

\noindent \textbf{Integrity}
Since the format in our task is strict and rigid, thus the number of words to be predicted is also pre-defined.
Our model must organize the language using the limited positions, thus sentence integrity may become a serious issue. For example, the integrity of ``love is not love . $\langle /s \rangle$'' is much better than``love is not the . $\langle /s \rangle$''. To conduct the evaluation of sentence integrity, we design a straightforward method by calculating the prediction probability of the punctuation characters before $\langle /s \rangle$ given the prefix tokens:
\begin{equation}
    Integrity = {2^{ - \frac{1}{{|Y|}}\sum\limits_{i = 1}^{|Y|} {\log (P(y_{punc}^i|y_0^i,y_1^i,...,y_{ < punc}^i))} }}
\label{eql:integrity}
\end{equation}
where $Y$ is the generated sequence of sentences. Smaller integrity metric value indicates higher sentence quality. To achieve this goal, we conduct pre-trainings for two GPT2 \cite{radford2019language} models on the large scale Chinese corpus and English corpus respectively. Then we utilize the GPT2 models to conduct the evaluation for sentence integrity.

\noindent
{\bf Human Evaluations}
For SongCi, we sampled 50 samples for 25 CiPais. For Sonnet, the whole 27 samples in the test set are selected for human evaluation. We recruit three helpers to score the \textit{Relevance}, \textit{Fluency}, and \textit{Style}. The rating criteria are as follows: \textit{Relevance}: \textbf{+2}: all the sentences are relevant to the same topic; \textbf{+1}: partial sentences are relevant; \textbf{0}: not relevant at all.  \textit{Fluency}: \textbf{+2}: fluent; \textbf{+1}: readable but with some grammar mistakes; \textbf{0}: unreadable. \textit{Style}: \textbf{+2}: match with SongCi or Sonnet genres; \textbf{+1}: partially match; \textbf{0}: mismatch.

\subsection{Comparison Methods}

\noindent \textbf{S2S} Sequence-to-sequence framework with attention mechanism \cite{bahdanau2014neural}. We regard the format and rhyme symbols $C$ as the input sequence, and the target as the output sequence. 

\noindent \textbf{GPT2}  We fine-tune the GPT2 models (the pre-training versions are used for sentence integrity evaluation) on SongCi and Sonnet respectively.

\noindent \textbf{SongNet} Out proposed framework with both the per-training and fine-tuning stages.

We also conduct ablation analysis to verify the performance of the defined symbols as well as the variants of model structures.

\begin{itemize}[topsep=0pt]
    \setlength\itemsep{-0.5em}
    \item \textbf{SongNet (only pre-tuning)} Without the fine-tuning stage.
    \item \textbf{SongNet (only fine-tuning)} Without the pre-training stage.
    \item \textbf{SongNet-GRU} Employ GRU \cite{cho2014learning} to replace Transformer as the core structure.
    \item \textbf{SongNet w/o C} Remove the format and rhyme symbols $C$.
    \item \textbf{SongNet w/o P} Remove the intra-position symbols $P$.
    \item \textbf{SongNet w/o S} Remove the sentence segment symbols $S$.
    \item \textbf{SongNet w/ inverse-P} Arrange the intra-position indices in ascending order instead of the descending order.
\end{itemize}

\begin{figure*}[!t]
    \centering
    \begin{minipage}{0.33\textwidth}
        \centering
        \includegraphics[width=0.9\textwidth]{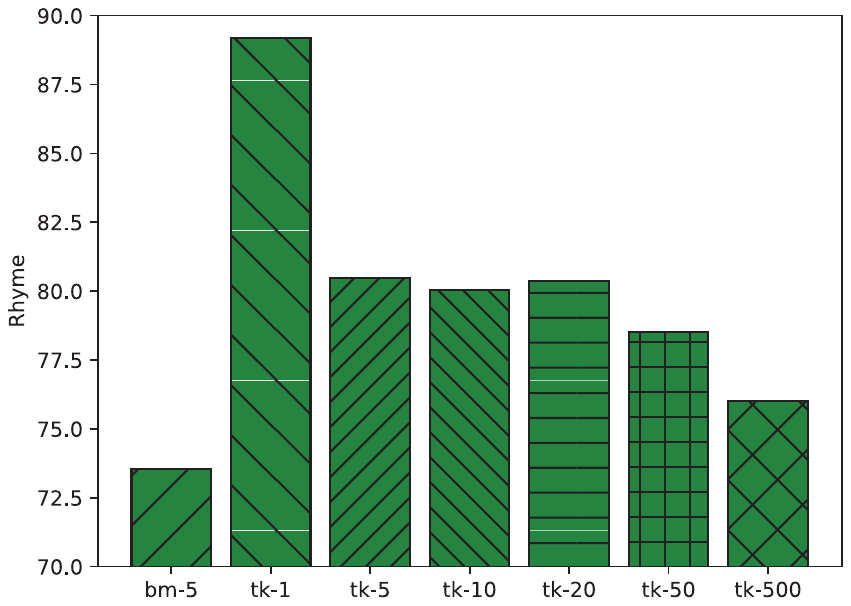} 
    \end{minipage}\hfill
    \begin{minipage}{0.33\textwidth}
        \centering
        \includegraphics[width=0.9\textwidth]{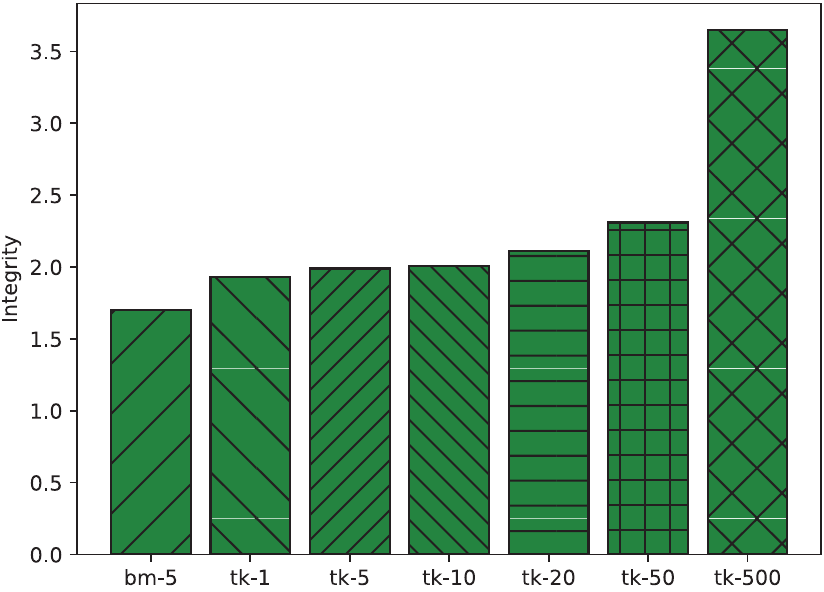} 
    \end{minipage}\hfill
    \begin{minipage}{0.33\textwidth}
        \centering
        \includegraphics[width=0.9\textwidth]{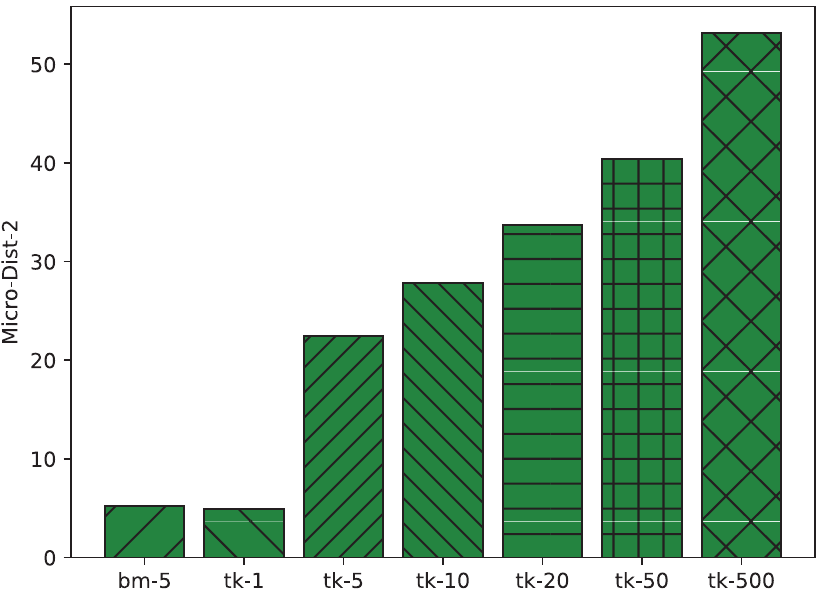} 
    \end{minipage}
    \caption{Parameter tuning of $k$ on the metrics of Rhyme, Integrity, and Micro-Dist-2.}
    \label{fig:k}
\end{figure*}

\begin{table*}[!t]
\centering
\includegraphics[width=2\columnwidth]{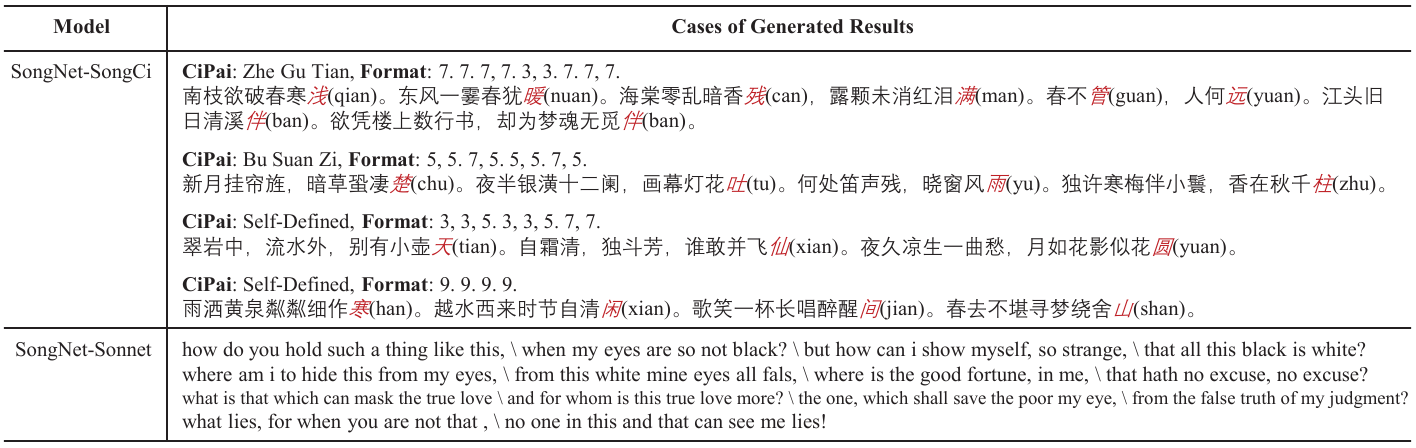}
\caption{Cases of the generated results for SongCi and Sonnet respectively. For SongCi, the number in Format (e.g., 3,5,7) denotes the number of tokens in one sentence. The rhyming words are labeled in red color and \textit{italic} font following is the Pinyin. (Since cases are provided to confirm the format consistency, thus we did not conduct translation for the Chinese samples. Translation for Chinese poetry is also a challenging task.)}
\label{tab:cases}
\end{table*}

\begin{table*}[!t]
	\centering
	\includegraphics[width=2\columnwidth]{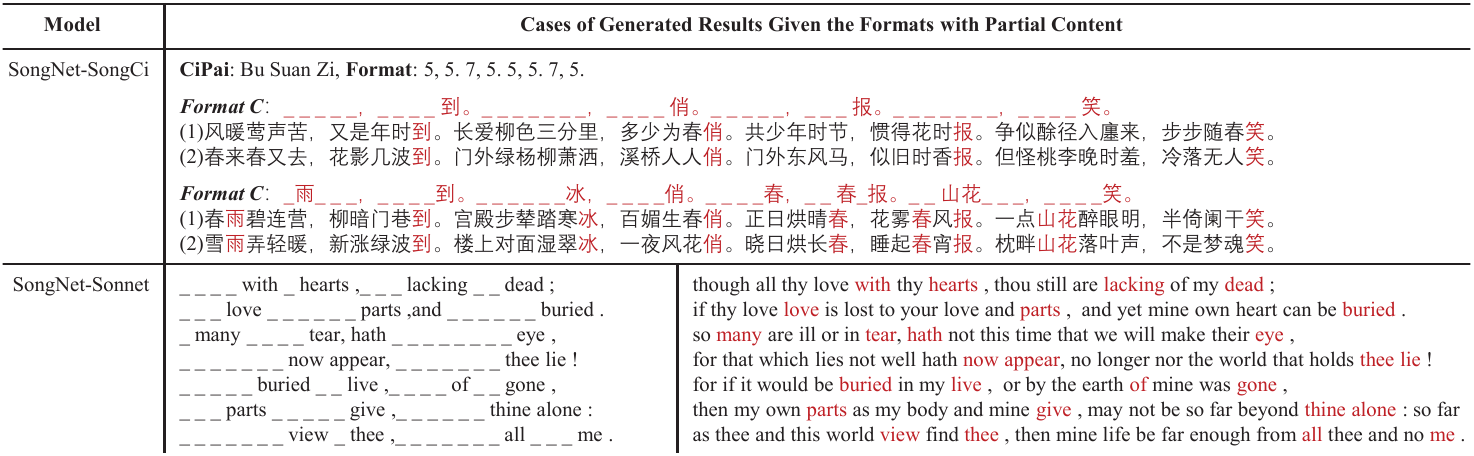}
	\caption{Cases of the generated results given the formats with partial pre-defined content. Format token ``\_'' needs to be translated to real word token.}
	\label{tab:polish}
\end{table*}

\section{Results and Discussions}

\subsection{Results}
Please note that we mainly employ top-$k$ sampling method \cite{fan2018hierarchical,radford2019language} to conduct the generation, and we let $k=32$ here. The parameter tuning of $k$ is described in Section~\ref{sec:tune}.

Table~\ref{tbl:exp-songci} and Table~\ref{tbl:exp-sonnet} depict the experimental results of SongNet as well as the baseline methods S2S and GPT2 on corpus SongCi and Sonnet respectively. It is obvious that our pre-training and fine-tuning framework SongNet obtain the best performance on most of the automatic metrics. Especially on the metric of Format accuracy, SongNet can even obtain a 98\%+ value which means that our framework can conduct the generation rigidly matching with the pre-defined formats.
On the metric of PPL, Rhyme accuracy, and sentence integrity, SongNet also performs significantly better in a large gap than the baseline methods such as S2S and GPT2 as well as the model variants only with the pre-training or fine-tuning stage. 

Another observation is that some of the results on corpus Sonnet are not as good as the results on SongCi. The main reason is that Sonnet only contains 100 samples in the training set as shown in Table~\ref{tab:datasets}. Therefore, the model cannot capture sufficient useful features especially for the rhyming issue.

\subsection{Ablation Analysis}

We conduct ablation study on corpus SongCi and the experimental results are depicted in Table~\ref{tbl:ablation-songci}. It should note that all the models are purely trained on SongCi corpus without any pre-training stages. From the results we can conclude that the introduced symbols $C$, $P$, and $S$ indeed play crucial roles in improving the overall performance especially on the metrics of format, rhyme, and sentence integrity. Even though some of the components can not improve the performance simultaneously on all the metrics, the combination of them can obtain the best performance.   

\subsection{Parameter Tuning}
\label{sec:tune}

Since we employ top-$k$ sampling as our main decoding strategy, thus we design several experiments to conduct the parameter tuning on $k$. We let k to be 1, 5, 10, 20, 50, 500 respectively. We also provide the beam-search (beam=5) results for comparing and reference.

The parameter tuning results are depicted in Figure~\ref{fig:k}. From the results we can observe that large $k$ can increase the diversity of the results significantly. But the Rhyme accuracy and the sentence integrity will drop simultaneously. Therefore, in the experiments we let $k=32$ to obtain a trade-off between the diversity and the general quality.

\subsection{Human Evaluation}

\begin{table}[!t]
\centering
\small
\resizebox{1\columnwidth}{!}{
\begin{tabular}{l|c|c|c}
\Xhline{2\arrayrulewidth}
 Model & Relevance & Fluency & Style\\
 \hline
 SongNet-SongCi & 1.36 & 1.45 & 2.00  \\
 \hline
 SongNet-Sonnet & 0.58 & 0.42 & 0.83    \\
 \Xhline{2\arrayrulewidth}
\end{tabular}}
\caption{Human evaluation results.}
\label{tab:human-eval}
\end{table}

For human evaluation, we just conduct the judging on the results generated by our final model SongNet. From the result we can observe that the results on corpus SongCi is much better than the ones on corpus Sonnet, which is because the corpus scale is different. And the the small scale also lead to dramatically dropping on all the metrics. 

\subsection{Case Analysis}
Table~\ref{tab:cases} depicts several generated cases for SongCi and Sonnet respectively. For SongCi, the formats (CiPai) are all cold-start samples which are not in the training set or even newly defined. Our model can still generate high quality results on the aspects of \textbf{format}, \textbf{rhyme} as well as \textbf{integrity}. However, for corpus Sonnet, even though the model can generate 14 lines text, the quality is not as good as SongCi due to the insufficient training-set (only 100 samples). \textit{We will address this interesting and challenging few-shot issue in the future}. 

In addition, we mentioned that our model has the ability of refining and polishing given the format $C$ which contains some fixed text information. The examples of the generated results under this setting are shown in Table~\ref{tab:polish}, which show that our model SongNet can generate satisfying results especially on SongCi.

\section{Conclusion}
We propose to tackle a challenging task called rigid formats controlled text generation. A pre-training and fine-tuning framework SongNet is designed to address the problem.
Sets of symbols are tailor-designed to improve the modeling performance for format, rhyme, and sentence integrity.
Extensive experiments conducted on two collected corpora demonstrate that our framework generates significantly better results in terms of both automatic metrics and  human evaluations given arbitrary cold start formats.

\bibliography{acl2020}

\begin{thebibliography}{27}
\expandafter\ifx\csname natexlab\endcsname\relax\def\natexlab#1{#1}\fi

\bibitem[{Bahdanau et~al.(2014)Bahdanau, Cho, and Bengio}]{bahdanau2014neural}
Dzmitry Bahdanau, Kyunghyun Cho, and Yoshua Bengio. 2014.
\newblock Neural machine translation by jointly learning to align and
  translate.
\newblock \emph{arXiv preprint arXiv:1409.0473}.

\bibitem[{Cho et~al.(2014)Cho, van Merrienboer, Gulcehre, Bahdanau, Bougares,
  Schwenk, and Bengio}]{cho2014learning}
Kyunghyun Cho, Bart van Merrienboer, Caglar Gulcehre, Dzmitry Bahdanau, Fethi
  Bougares, Holger Schwenk, and Yoshua Bengio. 2014.
\newblock Learning phrase representations using rnn encoder--decoder for
  statistical machine translation.
\newblock In \emph{Proceedings of the 2014 Conference on Empirical Methods in
  Natural Language Processing (EMNLP)}, pages 1724--1734.

\bibitem[{Dai et~al.(2019)Dai, Yang, Yang, Cohen, Carbonell, Le, and
  Salakhutdinov}]{dai2019transformer}
Zihang Dai, Zhilin Yang, Yiming Yang, William~W Cohen, Jaime Carbonell, Quoc~V
  Le, and Ruslan Salakhutdinov. 2019.
\newblock Transformer-xl: Attentive language models beyond a fixed-length
  context.
\newblock \emph{arXiv preprint arXiv:1901.02860}.

\bibitem[{Devlin et~al.(2019)Devlin, Chang, Lee, and
  Toutanova}]{devlin2019bert}
Jacob Devlin, Ming-Wei Chang, Kenton Lee, and Kristina Toutanova. 2019.
\newblock Bert: Pre-training of deep bidirectional transformers for language
  understanding.
\newblock In \emph{Proceedings of the 2019 Conference of the North American
  Chapter of the Association for Computational Linguistics: Human Language
  Technologies, Volume 1 (Long and Short Papers)}, pages 4171--4186.

\bibitem[{Fan et~al.(2018)Fan, Lewis, and Dauphin}]{fan2018hierarchical}
Angela Fan, Mike Lewis, and Yann Dauphin. 2018.
\newblock Hierarchical neural story generation.
\newblock In \emph{Proceedings of the 56th Annual Meeting of the Association
  for Computational Linguistics (Volume 1: Long Papers)}, pages 889--898.

\bibitem[{Gehring et~al.(2017)Gehring, Auli, Grangier, Yarats, and
  Dauphin}]{gehring2017convolutional}
Jonas Gehring, Michael Auli, David Grangier, Denis Yarats, and Yann~N Dauphin.
  2017.
\newblock Convolutional sequence to sequence learning.
\newblock In \emph{Proceedings of the 34th International Conference on Machine
  Learning-Volume 70}, pages 1243--1252. JMLR. org.

\bibitem[{Kingma and Ba(2014)}]{kingma2014adam}
Diederik~P Kingma and Jimmy Ba. 2014.
\newblock Adam: A method for stochastic optimization.
\newblock \emph{arXiv preprint arXiv:1412.6980}.

\bibitem[{Koehn(2004)}]{koehn2004pharaoh}
Philipp Koehn. 2004.
\newblock Pharaoh: a beam search decoder for phrase-based statistical machine
  translation models.
\newblock In \emph{Conference of the Association for Machine Translation in the
  Americas}, pages 115--124. Springer.

\bibitem[{Lau et~al.(2018)Lau, Cohn, Baldwin, Brooke, and
  Hammond}]{lau2018deep}
Jey~Han Lau, Trevor Cohn, Timothy Baldwin, Julian Brooke, and Adam Hammond.
  2018.
\newblock Deep-speare: A joint neural model of poetic language, meter and
  rhyme.
\newblock \emph{arXiv preprint arXiv:1807.03491}.

\bibitem[{Li et~al.(2016)Li, Galley, Brockett, Gao, and
  Dolan}]{li2016diversity}
Jiwei Li, Michel Galley, Chris Brockett, Jianfeng Gao, and Bill Dolan. 2016.
\newblock A diversity-promoting objective function for neural conversation
  models.
\newblock In \emph{Proceedings of the 2016 Conference of the North American
  Chapter of the Association for Computational Linguistics: Human Language
  Technologies}, pages 110--119.

\bibitem[{Li(2020)}]{li2020empirical}
Piji Li. 2020.
\newblock An empirical investigation of pre-trained transformer language models
  for open-domain dialogue generation.
\newblock \emph{arXiv preprint arXiv:2003.04195}.

\bibitem[{Li et~al.(2017)Li, Lam, Bing, and Wang}]{li2017deep}
Piji Li, Wai Lam, Lidong Bing, and Zihao Wang. 2017.
\newblock Deep recurrent generative decoder for abstractive text summarization.
\newblock In \emph{Proceedings of the 2017 Conference on Empirical Methods in
  Natural Language Processing}, pages 2091--2100.

\bibitem[{Liao et~al.(2019)Liao, Wang, Liu, and Jiang}]{liao2019gpt}
Yi~Liao, Yasheng Wang, Qun Liu, and Xin Jiang. 2019.
\newblock Gpt-based generation for classical chinese poetry.
\newblock \emph{arXiv preprint arXiv:1907.00151}.

\bibitem[{Radford et~al.(2018)Radford, Narasimhan, Salimans, and
  Sutskever}]{radford2018improving}
Alec Radford, Karthik Narasimhan, Tim Salimans, and Ilya Sutskever. 2018.
\newblock Improving language understanding with unsupervised learning.
\newblock Technical report, Technical report, OpenAI.

\bibitem[{Radford et~al.(2019)Radford, Wu, Child, Luan, Amodei, and
  Sutskever}]{radford2019language}
Alec Radford, Jeffrey Wu, Rewon Child, David Luan, Dario Amodei, and Ilya
  Sutskever. 2019.
\newblock Language models are unsupervised multitask learners.
\newblock \emph{OpenAI Blog}, 1(8).

\bibitem[{Rush et~al.(2015)Rush, Chopra, and Weston}]{rush2015neural}
Alexander~M Rush, Sumit Chopra, and Jason Weston. 2015.
\newblock A neural attention model for abstractive sentence summarization.
\newblock In \emph{Proceedings of the 2015 Conference on Empirical Methods in
  Natural Language Processing}, pages 379--389.

\bibitem[{See et~al.(2017)See, Liu, and Manning}]{see2017get}
Abigail See, Peter~J Liu, and Christopher~D Manning. 2017.
\newblock Get to the point: Summarization with pointer-generator networks.
\newblock In \emph{Proceedings of the 55th Annual Meeting of the Association
  for Computational Linguistics (Volume 1: Long Papers)}, pages 1073--1083.

\bibitem[{See et~al.(2019)See, Pappu, Saxena, Yerukola, and
  Manning}]{see2019massively}
Abigail See, Aneesh Pappu, Rohun Saxena, Akhila Yerukola, and Christopher~D
  Manning. 2019.
\newblock Do massively pretrained language models make better storytellers?
\newblock \emph{arXiv preprint arXiv:1909.10705}.

\bibitem[{Sennrich et~al.(2015)Sennrich, Haddow, and
  Birch}]{sennrich2015neural}
Rico Sennrich, Barry Haddow, and Alexandra Birch. 2015.
\newblock Neural machine translation of rare words with subword units.
\newblock \emph{arXiv preprint arXiv:1508.07909}.

\bibitem[{Shakespeare(2000)}]{shakespeare2000shakespeare}
William Shakespeare. 2000.
\newblock \emph{Shakespeare's sonnets}.
\newblock Yale University Press.

\bibitem[{Shang et~al.(2015)Shang, Lu, and Li}]{shang2015neural}
Lifeng Shang, Zhengdong Lu, and Hang Li. 2015.
\newblock Neural responding machine for short-text conversation.
\newblock In \emph{Proceedings of the 53rd Annual Meeting of the Association
  for Computational Linguistics and the 7th International Joint Conference on
  Natural Language Processing (Volume 1: Long Papers)}, pages 1577--1586.

\bibitem[{Speech@CMU(1998)}]{dictionary1998carnegie}
Speech@CMU. 1998.
\newblock Carnegie-mellon university pronouncing dictionary for american
  english.
\newblock \emph{Version 0.7b. Available at
  [http://www.speech.cs.cmu.edu/cgi-bin/cmudict]}.

\bibitem[{Vaswani et~al.(2017)Vaswani, Shazeer, Parmar, Uszkoreit, Jones,
  Gomez, Kaiser, and Polosukhin}]{vaswani2017attention}
Ashish Vaswani, Noam Shazeer, Niki Parmar, Jakob Uszkoreit, Llion Jones,
  Aidan~N Gomez, {\L}ukasz Kaiser, and Illia Polosukhin. 2017.
\newblock Attention is all you need.
\newblock In \emph{Advances in neural information processing systems}, pages
  5998--6008.

\bibitem[{Vinyals and Le(2015)}]{vinyals2015neural}
Oriol Vinyals and Quoc Le. 2015.
\newblock A neural conversational model.
\newblock \emph{arXiv preprint arXiv:1506.05869}.

\bibitem[{Yang et~al.(2019)Yang, Dai, Yang, Carbonell, Salakhutdinov, and
  Le}]{yang2019xlnet}
Zhilin Yang, Zihang Dai, Yiming Yang, Jaime Carbonell, Ruslan Salakhutdinov,
  and Quoc~V Le. 2019.
\newblock Xlnet: Generalized autoregressive pretraining for language
  understanding.
\newblock \emph{arXiv preprint arXiv:1906.08237}.

\bibitem[{Zhang and Lapata(2014)}]{zhang2014chinese}
Xingxing Zhang and Mirella Lapata. 2014.
\newblock Chinese poetry generation with recurrent neural networks.
\newblock In \emph{Proceedings of the 2014 Conference on Empirical Methods in
  Natural Language Processing (EMNLP)}, pages 670--680.

\bibitem[{Zhu et~al.(2015)Zhu, Kiros, Zemel, Salakhutdinov, Urtasun, Torralba,
  and Fidler}]{zhu2015aligning}
Yukun Zhu, Ryan Kiros, Rich Zemel, Ruslan Salakhutdinov, Raquel Urtasun,
  Antonio Torralba, and Sanja Fidler. 2015.
\newblock Aligning books and movies: Towards story-like visual explanations by
  watching movies and reading books.
\newblock In \emph{Proceedings of the IEEE international conference on computer
  vision}, pages 19--27.

\end{thebibliography}
\bibliographystyle{acl_natbib}


\end{document}